# Part-based Multi-stream Model for Vehicle Searching

Ya Sun, Minxian Li *, Jianfeng Lu
School of Computer Science and Engineering,
Nanjing University of Science and Technology,
Nanjing, China
Email: { minxianli, sunya, lujf } @njust.edu.cn

*Abstract*—Due to the enormous requirement in public security and intelligent transportation system, searching an identical vehicle has become more and more important. Current studies usually treat vehicle as an integral object and then train a distance metric to measure the similarity among vehicles. However, these raw images may be exactly similar to ones with different identification and include some pixels in background that may disturb the distance metric learning. In this paper, we propose a novel and useful method to segment an original vehicle image into several discriminative foreground parts, and these parts consist of some fine grained regions that are named discriminative patches. After that, these parts combined with the raw image are fed into the proposed deep learning network. We can easily measure the similarity of two vehicle images by computing the Euclidean distance of the features from FC layer. Two main contributions of this paper are as follows. Firstly, a method is proposed to estimate if a patch in a raw vehicle image is discriminative or not. Secondly, a new Part-based Multi-Stream Model (PMSM) is designed and optimized for vehicle retrieval and re-identification tasks. We evaluate the proposed method on the VehicleID dataset, and the experimental results show that our method can outperform the baseline.

## I. Introduction

Recent years, due to the widely distributed cameras and the improvement of the computer vision technology and computing capability, the requirement of specific vehicle retrieval and re-identification has grown rapidly in some areas such as public security and intelligent transportation, and it has attracted more and more attentions. Although the license plate recognition has been widely applied to traditional transportation system, it fails for some specific scenario such as occlusion or invisible license plate.

Thus, some other methods using more information of vehicle object provide an alternative. For example, at the beginning, a simple deep classification model is used to classify the identities [1]. However, most of the concrete applications that aim at searching a specific vehicle are lack of training data. And deep classification model do not work well under only two to three samples for one identity. Then, Siamese structure based methods that use image pairs or triplets as input have been used commonly and achieved competitive performance in many vision tasks such as face recognition [2][3], fine-grained classification [4][5], person re-identification [6][7] and vehicle re-identification [8][9][10] as well. Above methods are suitable for dealing with the datasets that have a large number of categories (e.g., thousands of flowers, countless faces) and have few samples in most of categories.

Another common problem is how to classify exactly with a large of categories and these categories may have high inter-class similarity and high intra-class difference. Luckily, the visual features on some particular regions are unique and obvious, these features can help human to recognize a vehicle from coarse level to fine level. For example, two vehicles may have the same make in coarse level, but models and identifications are different in medium level and fine level respectively. In human's experience, firstly, some common elements on a vehicle like the shape of the headlamps help to recognize different models and then the decorations may help to recognize different identifications (see Fig. 1). However, these specific semantic parts are unstable or even invisible sometimes. What we have done in this paper is how to find some non-semantic parts and to integrate these parts as the input to a Siamese structure which is adept at training a good distance metric. Finally, we propose our part-based method to improve the accuracy of re-identifying a vehicle.

In this paper, there are two contributions: (1) We design a method to score a fine-grained patch in different levels by the vehicle hierarchy labels. We then gather the patches with the highest score to form a larger part that can discriminate different vehicle at different classification grade like make, type, identification. (2) Furthermore, a multi-stream deep learning network that takes the discriminative parts along with the raw image as input and learns the part-based representation of a vehicle. Specifically, at the end of the network, we optimize a triplet loss which doesn't need any category labels. And that similarity constraint is proved to be suitable for scenario that have numerous categories in test set and some in which are unseen in training set, such as vehicle retrieval and vehicle re-identification tasks.

The rest of the paper is organized as follows: Section 2 provides a brief review about part-based fine-grained image classification methods and the representative methods for vehicle re-identification or retrieval recent years. Section 3 discusses the proposed discriminative part mining method and the unified deep learning network. Experimental results and analysis is presented in Section 4. At last, in Section 5, we make a conclusion of this paper.

*Corresponding author: Minxian Li (email: minxianli@njust.edu.cn).

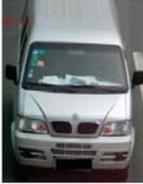

Fig. 1. There are several parts which are discriminative and helpful to recognize a specific vehicle step by step(make, model, identification). In this paper, we propose a mining method to collect some parts which are located in nearly the same positions with manually-labeled ones.

## II. RELATED WORK

The pedestrian in surveillance videos is the main target which in most of the previous researches about the object identification. Before the deep learning was used into person re-ID [6][11], the most commonly used features are the color and texture features [12][13]. Also, researchers chose some mid-level representations by well artificially-designed to replace the low-level features or be as a supplement [14].

Hand-crafted re-ID systems are always relied on a good kind of representation as well as a good distance metric which make the representation invariant to some visually different images from an identity. Luckily, these two steps are unified into one step in CNN based deep learning system. In summary, two common methods based on a CNN model are employed in the person re-identification community. The first one uses a classification model which is a simple CNN network with a commonly used softmax loss at the end [15] and then all the channels from a medium layer of the net are extracted to form the representation whose length is equal to the number of the channels'.

The second type of CNN model is more prevalent up till now which is called the Siamese Structure [16], which is suitable for the scenario of lack of training data and only two or three images available for one class or identity in most of re-ID datasets. The reason that Siamese model performs well for these situations is that it does not use the true category labels which is not sufficient to guide the training but use "relationship labels" between two or three input images like "A and B are same class" or "A and B do not belong to the same identity". An instance that uses the Siamese model in the field of person re-identification is [6]. In this paper, two person images are separated into three overlapped parts (heads ,bodies, legs) and then these parts go through three basic Siamese models in pairs whose output eventually is the similarity of the two images.

The most advanced model comparing to the basic Siamese Network is the triplet loss structure which take three images as input and the constraint is "A is more similar to B than C" [3]. Many similar structures are proposed the last two years whose target is not the pedestrian but the vehicles [8][9][10]. Deep Relative Distance Learning method was proposed by Liu et al. [8], this structure takes two clusters instead of three single images as input and the triplet loss function was replaced by a new loss function named 'Coupled Cluster Loss'. This structure will not be so sensitive to the selection of one triplet sample. Zhang et al. [10] designed a deep structure to take four images which have the hierarchy labels as input and optimized with a new constraint of these four images. The representation based on such structure not only can include the fine-grained information but also the coarse information, which means more hierarchical information can be provided.

Some researchers have proposed that the fine-grained sub-categories share the same parts and can be discriminated by these subtle parts (e.g., different wings on birds, different eyes on cats) [17][18][19][20][21]. Some of these methods rely on manually-labeled strong part annotations. For example, Huang et al. used a network architecture called the Part-Stacked CNN which use a part stream in which multiple parts located by a FCN stream and share the same feature extraction procedure. Then these features are linked with additional feature extracted by the third stream which uses the original image of lower spatial-resolution as input [17]. And the others do not need part annotations. Such as in [21], Wang et al. propose a method that can automatically mine some discriminative geometrically-constrained triplets which is a kind of middle level descriptor for vehicle classification. Inspired by these ideas, we propose our approach that the raw image and the parts and are treated as input respectively and then thrown into our multi-stream deep learning network.

In this paper, we firstly propose how to mine some discriminative patches on vehicles by a hierarchy labels (e.g. model and identification), and these patches are practically gathered into two fixed areas so called the discriminative parts. We then propose our new deep learning network to use three streams to embed the two parts and the raw image.

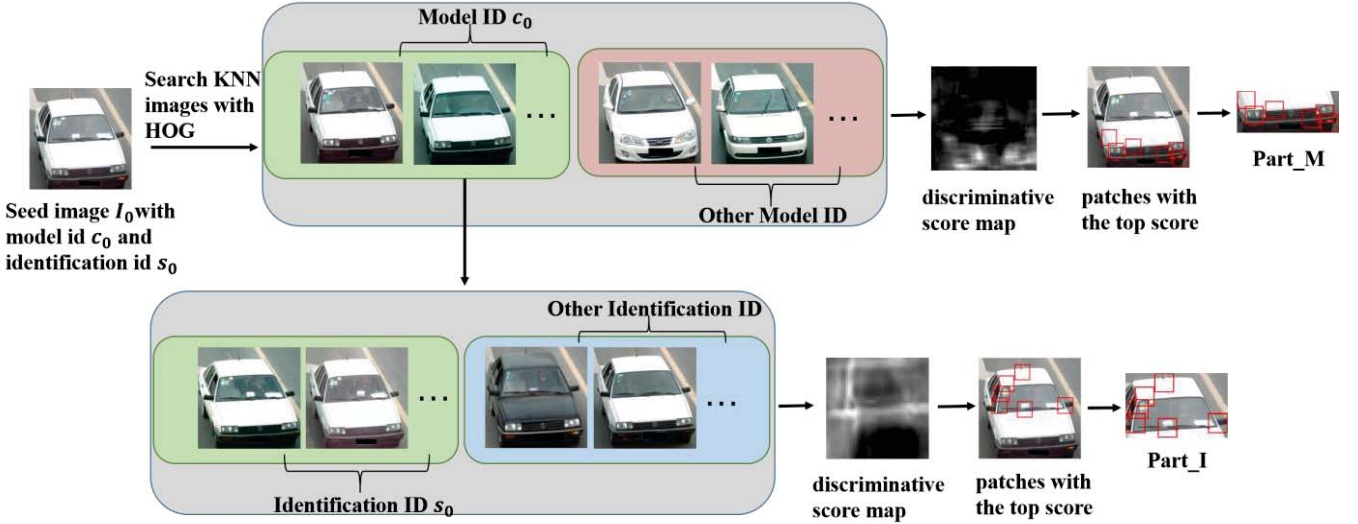

Fig. 2. The flowchart of mining discriminative parts on different levels.

## III. METHODOLOGY

### A. Discriminative Parts Mining

As far as we know, it is the first time that using vehicle parts for vehicle re-ID and vehicle retrieval tasks. In this section, we propose a method that mines the patches on a vehicle and show that why these patches can help to discriminate a vehicle at different classification grade like vehicle models or vehicle identities.

Given a vehicle training set $I$, $C$ is the set of the vehicle model labels and $C = \{c_0, c_1, c_2, ..., c_P\}$, $S$ is the set of the vehicle identification labels and $S = \{s_0, s_1, s_2, ..., s_Q\}$. Besides, $\|\cdot\|$ means to compute the Euclidean distance.

Inspired by the study [21], given a seed training image $I_0$ with class label $c_0$ and identification label $s_0$, we firstly search k-nearest neighbors (KNN) images by HOG feature in the training set. After collecting images $\{I_i\}_{i=1}^{M}$, we then compute the discriminative score at position $(x, y)$ by patches in the same position of each of these neighbor images in Eq. (1) and aim to find such a position $(x, y)$ that maximize the ratio of inter-class difference and intra-class difference.

$$d(x,y) = \frac{\sum_{c=c_1}^{c_P} \left\| \overline{F_c}(x,y) - \overline{F}(x,y) \right\|}{\sum_{c=c_1}^{c_P} \sum_{I_i \in c} \left\| F_{I_i}(x,y) - \overline{F_c}(x,y) \right\|} \quad (1)$$

where $\overline{F_c}(x, y)$ is average feature of patches from class $c$, $\overline{F}(x, y)$ is average feature of all the patches. $I_i \in c$ means that vehicle image $I_i$ has identification label $c$ and its feature is $F_{I_i}(x, y)$.

However, it is unnecessary to obtain a patch that can recognize all of the vehicle models. We just need to discriminate the class of seed image $c_0$ with the others. So we redefine the discriminative score function as follow:

$$d(x,y) = \frac{\min_{c \in C, c \neq c_0} \left\| \overline{F_c}(x,y) - \overline{F_{c_0}}(x,y) \right\|}{\max_{I_i, I_0 \in c_0} \left\| F_{I_i}(x,y) - F_{I_0}(x,y) \right\|} \quad (2)$$

where inter-class difference is replaced by the distance between class $c_0$ and the nearest class, the intra-class difference is replaced by a distance between $I_0$ and the furthest one of class $c_0$.

In practice, we select 6 patches with the top discriminative score as the discriminative patches in the seed image as in [21], and then define a minimum bounding rectangle that contain these 6 patches called the discriminative part. The part is always related to a fixed area on the vehicle according to our observation. We also find the second discriminative part which fixed in another area using the identification label $s_0$ of seed image in Eq. (3).

$$d(x,y) = \frac{\min_{s \in c_0, s \neq s_0} \left\| \overline{F_s}(x,y) - \overline{F_{s_0}} \right\|}{\max_{I_i, I_0 \in s_0} \left\| F_{I_i}(x,y) - F_{I_0}(x,y) \right\|} \quad (3)$$

where $s_0$ means the identification label of seed image and the $s \in c_0$ means all the vehicle identities with identification labels $s$ that belong to the same vehicle model with model label $c_0$.

Our method is very similar to humans to identify a vehicle in several steps: firstly identify the coarser level like model of a vehicle by the "lower face", secondly identify the finer level like the specific vehicle belong to someone by some decorates in the windshield. We think these two parts help searching identical vehicle in vehicle re-ID and vehicle retrieval tasks.

We finally name the discriminative part which can help to discriminate vehicle models as **Part_M** and name the part which can help to discriminate vehicle identity as **Part_I**. We sample these two types of discriminative parts by using the proposed method and then train a detector (we use SSD [22] to train a detector in our experiment) to locate these two parts in our deep learning network.

### B. Triplet Loss Network

Next, we use a standard triplet loss to make the input triplet units $<x^a, x^p, x^n>$ satisfying the following constraint:

$$\|f(x^a) - f(x^p)\| + \alpha \le \|f(x^a) - f(x^n)\| \quad (4)$$

where $x^a$ means to select an image as anchor, $x^p$ means a positive image having the same identification with the anchor, and $x^n$ means a negative one having the different identification with the anchor. The hyper-parameter $\alpha$ measures a margin between the intra-class distance and the inter-class distance.

In other word, the triplet loss minimizes the distance between the anchor and positive one and maximizes the distance between the anchor and negative one. The loss function is defined as follow:

$$L = \sum_i^N \left[ \|f(x_i^a) - f(x_i^p)\| - \|f(x_i^a) - f(x_i^n)\| + \alpha \right]_+ \quad (5)$$

Especially, we take N triplets in a batch to decrease the influence taken from some improper triplets and use a moderate triplets mining strategy used in [7] to accelerate the convergence speed.

### C. Part-based 3-stream model

As the name suggests, the complete deep learning network we proposed consists of three separate CNN streams as shown in Fig. 3. And these CNN streams take Part_M, Part_I and the normalized original images as inputs. More specifically, these streams have the same CNN architecture but do not share parameters.

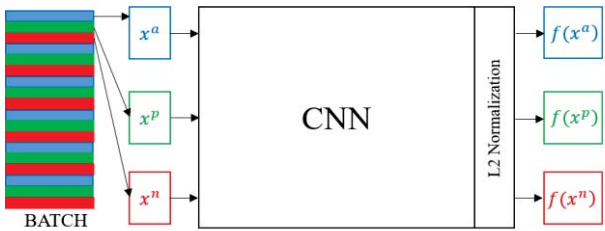

Fig. 3. The flowchart is a standard triplet loss network without the triplet loss function. We take it as one stream of the total three streams in our unified network. For example, in "Part_I" stream, it takes three "Part_I" part images as input each time.

One reasonable explanation to use the multi-stream architecture is that the parts cropped from the original image have a larger scale than the original image when they share the same size and multi-scale information is helpful to improve the performance of the image matching. Besides, we consider these two discriminative parts twice which are the most powerful to discriminate vehicles of different models or different identifications.

Actually, we can use any CNN architecture and the complete flow consists of following steps: (1) At first, we use 3N original images in a batch to form N inputs, and each input is a triplet and consists of anchor image, positive image and negative image. There are 3 type of input then to 3 streams respectively, namely "Part_M" stream, "Part_I" stream and "Whole image" stream, the detail can be referred to Fig. 4. Inputs in each stream are in order (anchor, positive, negative, anchor…) and the input images have the same size. (2) Then, we get 9N vectors if all these streams end with a FC layer and then we concatenate 3 vectors coming from 3 different streams each times. Now, we get 3N features again whose length is set to 1024 by a specific FC layer. (3) Finally, the triplet loss process 3 features at a time and we obtain N loss value. Specially, in testing stage, we use the 1024-dimension feature from the FC layer as described in Fig. 4 to represent a vehicle whose length is the same as which in baseline.

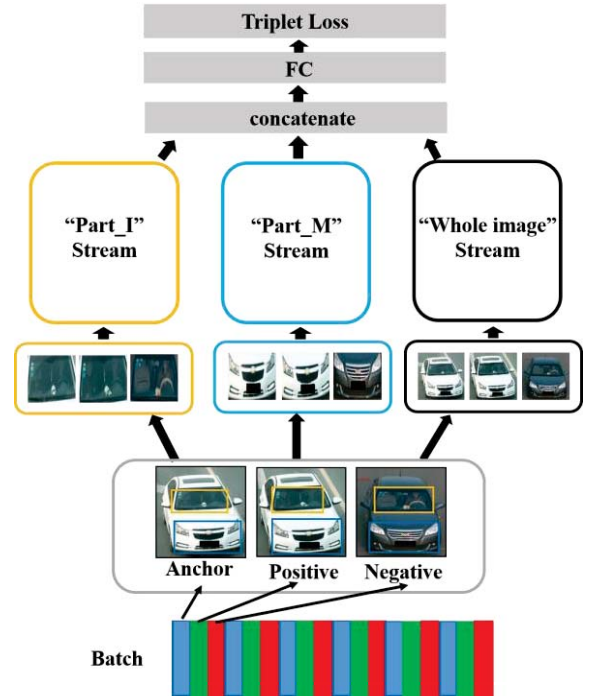

Fig. 4. The flowchart of our proposed multi-stream deep learning network

## IV. EXPERIMENTS

To evaluate the performance of our method, we use the dataset "VehicleID" [8] which is lately widely used in vehicle vision tasks, and, this dataset is large-scale and captured from surveillance videos. In training set, there are 113,346 images of 13,164 vehicle identities with unique vehicle id which are used to train our deep learning network. Besides, another label "model id" is combined with the "vehicle id" label to extract the discriminative part samples.

The baseline in this paper is the DRDL structure with mixed difference in [8]. Following some settings in their paper, the test set is split into three data sets which is demonstrated in TABLE I. Two main tasks in vehicle visual problems (Vehicle Retrieval and Vehicle Re-identification) are addressed in our paper and all experimented in these three test datasets. For simply comparison with the baseline, we firstly use the "VGG_CNN_M_2048" network and then use a more effective but not deeper network which makes great progress.

We use deep learning framework Caffe to train our part-based multi-stream model. In training process, we set the momentum $\mu$ to 0.9, weight decay $\lambda=2\times10^{-4}$, base learning rate $\eta^{(0)}$ to 0.05 and learning rate $\eta^{(i)} = \eta^{(0)} \times 0.9^{\lfloor n/10000 \rfloor}$ (n is the iteration number). Batch size in the training is set to 180. Specifically, we set the original margin α in triplet loss to 0.1 and $\alpha = 0.1 \times \lceil n/10000 \rceil$ and the maximum iteration number is set to 100k.

TABLE I. THREE TEST SETS WITH DIFFERENT SCALE

|  | Small | Medium | Large |
|---|---|---|---|
| Vehicle Identities | 800 | 1600 | 2400 |
| Image Numbers | 6493 | 13377 | 19777 |

### A. Vehicle Retrieval

For vehicle retrieval task, we use the mean average precision (mAP) as evaluation metric. In detail, we randomly select one image for all the vehicle images with the same identification. We then obtain 800, 1600, 2400 images in probe sets and the rest are throw into the gallery sets. After extracting the feature from our deep learning network, we can simply measure the similarity between two vehicle images by their Euclidean distance. The results on three test sets are shown in TABLE II.

TABLE II. MAP OF DIFFERENT METHODS IN RETRIEVAL TASK

| mAP | Small | Medium | Large |
|---|---|---|---|
| VGG+Triplet Loss | 0.444 | 0.391 | 0.373 |
| VGG+CCL | 0.492 | 0.448 | 0.386 |
| MixedDiff_CCL | 0.546 | 0.481 | 0.455 |
| VGG+PMSM | **0.567** | **0.532** | **0.495** |

In TABLE II. the first two deep structures are proposed in baseline. The first one uses the VGG_CNN_M_2048 with the Couple Cluster Loss and the second with mixed difference includes with two Couple Cluster Loss and several Softmax Loss. However, our method using part-based representation outperforms them on all the three test sets.

### B. Vehicle Re-identification

Another task is vehicle re-identification which is closely related to public security area. In this task, we use the match rate at top-1 and top-5 as evaluation metric. We exchange the probe sets and the gallery sets in vehicle retrieval task. In other words there are N images from N vehicle identifications in gallery sets and the rest from all the vehicle identifications are put into the probe sets. The results are shown in TABLE III.

TABLE III. MATCH RATE AT TOP-1&5 IN RE-IDENTIFICATION TASK

| Match Rate | | Small | Medium | Large |
|---|---|---|---|---|
| VGG+Triplet Loss | Top1 | 0.404 | 0.354 | 0.319 |
| VGG+CCL | Top1 | 0.436 | 0.370 | 0.329 |
| MixedDiff_CCL | Top1 | 0.490 | 0.428 | 0.382 |
| VGG+PMSM | Top1 | **0.568** | **0.548** | **0.495** |
| VGG+Triplet Loss | Top5 | 0.617 | 0.546 | 0.503 |
| VGG+CCL | Top5 | 0.642 | 0.571 | 0.533 |
| MixedDiff_CCL | Top5 | **0.735** | 0.668 | 0.616 |
| VGG+PMSM | Top5 | 0.707 | **0.695** | **0.662** |

It can be seen in TABLE III. that all the results from our method are have an improvement compared with the baseline except for one exception. However, the match rate at top-1 is more important in reality and dataset become larger and larger which have to search nowadays. So our method whose score has a great increase comparing with the baseline in top1 and is more stable with the expanding dataset.

### C. Better Deep Network

Deep residual learning is proved to be an efficiency method to ease the vanishing gradient problem when training a deeper and deeper neural network. But in our paper, this structure makes the representation of a vehicle object include more information from lower-layer which may be helpful to recognize a vehicle. In our experiment, a simple residual network called "resnet-18" proposed by He et al. [23] is used for vehicle retrieval and vehicle re-identification tasks.

TABLE IV. MAP OF OUR METHOD WITH VGG AND RESNET-18

| mAP | Small | Medium | Large |
|---|---|---|---|
| VGG+PMSM | 0.567 | 0.532 | 0.495 |
| Resnet+PMSM | **0.642** | **0.572** | **0.518** |

For vehicle retrieval task, the results are shown in TABLE IV. In conclusion, our 3-stream deep learning network with resnet-18 beats the one with VGG network nearly 5% on the three test datasets. And for the vehicle re-identification task, the results are in TABLE V. The result shows that the resnet-18 has a great improvement on our method with VGG network.

TABLE V.   MATCH RATE OF OUR METHOD WITH VGG AND RESNET-18

| Match Rate | | Small | Medium | Large |
|---|---|---|---|---|
| VGG+PMSM | Top1 | 0.568 | 0.548 | 0.495 |
| Resnet+PMSM | | **0.622** | **0.5549** | **0.5047** |
| VGG+PMSM | Top5 | 0.707 | 0.695 | 0.662 |
| Resnet+PMSM | | **0.845** | **0.7958** | **0.7574** |

## V. CONCLUSION

In this paper, a part-based multi-stream model (PMSM) is proposed for vehicle re-identification and vehicle retrieval tasks. We use a discriminative parts mining method to find two fixed parts that can help distinguish vehicles at different classification grade. Then we use three streams to accommodate these parts with a similarity constraint in order to get a more discriminative representation of a vehicle. Although much more researches on vehicle aim to find a more powerful similarity constraint between vehicles nowadays, we propose a new method that uses more discriminative parts of a vehicle and the experiments prove that this method is efficient and promising.

ACKNOWLEDGMENT

This work is partially supported by National Natural Science Foundation of China (Grant No. 61401212), and Technology Support Program of Jiangsu Province (Grant No. BE2015728).